\documentclass[runningheads]{llncs}
\usepackage[T1]{fontenc}
\usepackage{graphicx}
\usepackage{amsmath}
\usepackage{amssymb}
\usepackage{booktabs} %for table
\usepackage{tabularx}%for table
\begin{document}

\title{Is Uncertainty Quantification a Viable Alternative to Learned Deferral?}
\titlerunning{Is UQ a Viable Alternative to Learned Deferral?}

\author{Anna M. Wundram\inst{1} \and Christian F. Baumgartner \inst{1,2}}

\authorrunning{A. M. Wundram and C. F. Baumgartner}

\institute{Faculty of Health Sciences and Medicine, University of Lucerne, Switzerland
\email{firstname.lastname@unilu.ch}
\and
Cluster of Excellence -- ML for Science, University of Tübingen, Germany
}

\maketitle              

\begin{abstract}
\setcounter{footnote}{0} 
Artificial Intelligence (AI) holds the potential to dramatically improve patient care. However, it is not infallible, necessitating human-AI-collaboration to ensure safe implementation. One aspect of AI safety is the models' ability to defer decisions to a human expert when they are likely to misclassify autonomously. Recent research has focused on methods that \textit{learn} to defer by optimising a surrogate loss function that finds the optimal trade-off between predicting a class label or deferring. 
However, during clinical translation, models often face challenges such as data shift. Uncertainty quantification methods aim to estimate a model’s confidence in its predictions. However, they may also be used as a deferral strategy which does not rely on learning from specific training distribution. We hypothesise that models developed to quantify uncertainty are more robust to out-of-distribution (OOD) input than learned deferral models that have been trained in a supervised fashion.
To investigate this hypothesis, we constructed an extensive evaluation study on a large ophthalmology dataset, examining both learned deferral models and established uncertainty quantification methods, assessing their performance in- and out-of-distribution. Specifically, we evaluate their ability to accurately classify glaucoma from fundus images while deferring cases with a high likelihood of error.
We find that uncertainty quantification methods may be a promising choice for AI deferral\footnote{Code available at https://github.com/annawundram/UQforDeferral }.
\end{abstract}
\section{Introduction}
Artificial Intelligence (AI) plays a promising role in many medical fields including ophthalmology, holding the potential to dramatically improve patient care through reduced waiting times or improved diagnostic accuracy~\cite{beede2020human,ruamviboonsuk2022real}. Nonetheless, AI models are not infallible and including AI without any human supervision might lead to fatal mistakes. A promising paradigm to ensure clinical safety are deferral mechanisms, where
the AI model functions autonomously in the majority of cases deferring only uncertain cases to a human expert~\cite{heydon2021prospective,macdonald2025generating}.

\begin{figure}[t]
    \centering
    \includegraphics[width=\textwidth]{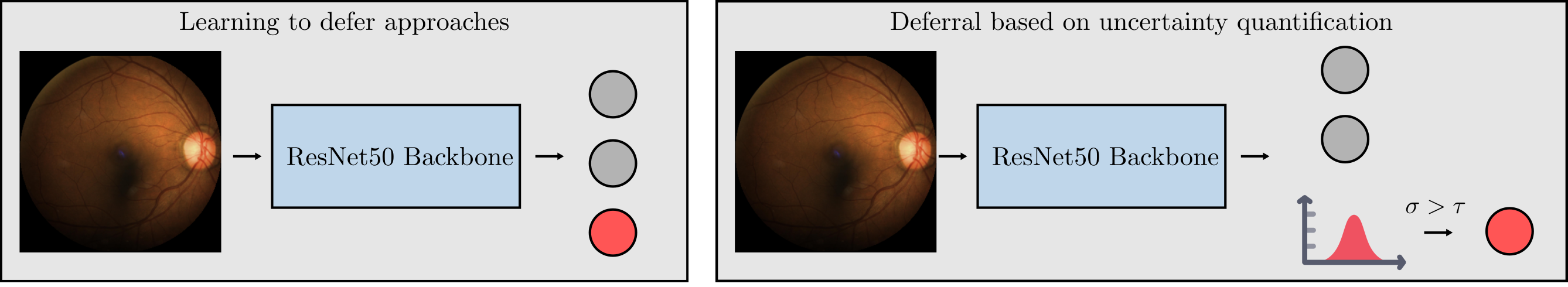}
    \caption{\textbf{Overview}. We compare two AI deferral paradigms. Learning to defer (left) adds a deferral class (red) and jointly optimizes prediction and deferral. Uncertainty-based (right) estimates uncertainty independently and defers based on a threshold.}
    \label{fig:teaser}
\end{figure}

A widely explored paradigm in AI deferral is \textit{learned deferral}, in which classifiers are trained with an option to defer the decision to an expert rather than making a prediction. In this paradigm, the classifier and deferral model can either be trained jointly~\cite{hemmer2022forming,keswani2021towards,mao2024predictor} or as a two-stage approach~\cite{alves2024cost,de2024towards,mao2024two,zhang2025learning}. In both cases, most prior methods extend the label space by an additional deferral class. The learned deferral model is optimised to predict a class label when the probability is high that it will make a correct prediction and to predict the deferral class when unsure (see Fig.~\ref{fig:teaser} left). 
%In the latter case, the final decision is then deferred to a human expert. 
This model behaviour is achieved by optimising a surrogate loss, which leads to a trade-off between the model making an autonomous decision or deferral. Mozannar et al.~\cite{mozannar2020consistent} were the first to introduce a Bayes-consistent surrogate loss by generalising the cross-entropy loss for cost-sensitive learning. Multiple papers have subsequently been published refining this loss by ensuring calibration~\cite{verma2022calibrated}, stronger consistency guarantees~\cite{mao2024predictor,mao2024realizable,mozannar2023should}, and allowing for more generalised cost functions~\cite{mao2024principled,mao2024realizable}. 

A related field is uncertainty quantification, which aims to obtain a measure of uncertainty of a model's prediction~\cite{lambert2022trustworthy}. Uncertainty estimates can also inform deferral decisions by identifying inputs with uncertainty above a specified threshold and routing them to a human expert (see Fig.~\ref{fig:teaser} right). While some techniques use deferral settings as a qualitative evaluation strategy for uncertainty quantification techniques~\cite{lambert2022trustworthy}, their explicit utility for deferral is understudied. In particular, it is unknown how uncertainty quantification methods compare to the more widely studied learned deferral methodology. Only a few approaches employ uncertainty specifically for deferral. Dvijotham et al.~\cite{dvijotham2023enhancing} proposed an approach that tunes a range of the model's confidence score within which the diagnosis should be deferred. Fang et al.~\cite{fang2024learning} create conformal sets based on model's confidence which represent its uncertainty about whether to defer or not.  Lastly, Liu et al.~\cite{liu2022incorporating} proposed a two-stage approach that consists of an ensemble and a deferral model using the ensemble's uncertainty. 

For deferral methods to be effective in clinical practice, they must be robust to challenges commonly faced during clinical translation such as OOD inputs caused by data shifts. We hypothesize that models developed to quantify uncertainty are more robust to OOD input than learned deferral models that have been trained in a supervised fashion, as they are inherently designed to estimate uncertainty rather than learn deferral behaviour from domain-specific patterns.

To assess the feasibility of uncertainty quantification techniques for AI deferral, we constructed an extensive evaluation study that examines learned deferral models as well as various established uncertainty quantification approaches in in- and out-of-domain settings. We show that uncertainty quantification methods indeed prove to be a promising choice for learning to defer.

\section{Methods}
\label{sec:methods}

In this study, we consider the general setting of a binary medical image classification system with a deferral mechanism. We compare two learned deferral approaches and five approaches based on uncertainty in their ability to correctly defer images that are likely to be misclassified. We perform our study on the diagnosis of glaucoma from fundus images.  Glaucoma is a disease of the eye characterised by a gradual degeneration of nerve fibres and is the second leading cause of blindness worldwide. AI assisted diagnosis of glaucoma may be a cost- and time-effective solution to screen for the disease in large populations~\cite{de2023airogs}.

\subsection{Compared Deferral Methods}
\label{sec:compared_def_meth}
We compare two main categories of deferral approaches: (1) learned deferral, where the classifier is trained with an additional deferral class using a dedicated loss function and (2) uncertainty-based deferral, where deferral decisions are made by thresholding uncertainty estimates. To ensure comparability, all methods utilise a ResNet50 architecture~\cite{he2016deep}  as their backbone.

\subsubsection{Learned Deferral Methods}
\label{sec:def_learned}
In a regular classification problem, a model $m$ predicts a label $\hat{y}$ from the label space ${\mathcal{Y}} = \{1, \dots n\}$ 
given an input $x$ and target $y$. For learned deferral, this label space is extended to $\bar{\mathcal{Y}} = \{1, \dots n, n+1\}$ now including $n+1$ as the "deferral class". Predicting this class defers the decision to a human expert, which incurs a cost. Finding the optimal trade-off between autonomous decision and deferral is achieved by optimising a cost-sensitive surrogate loss. 

\textbf{Learned deferral (one-stage)} -- The first method we examine in this category, is the one-stage deferral method  proposed by Mao et al.~\cite{mao2024realizable}. This method is trained using the following realisable $\mathcal{H}$-consistent and Bayes-consistent loss, which has been derived from the logistic loss
\begin{equation}
   l(m, x, y) = -\alpha~log\biggl[\dfrac{e^{m(x,y)}}{\sum_{y'\in\bar{\mathcal{Y}}}e^{m(x,y')}}\biggr] - (1-\alpha)~log\biggl[\dfrac{e^{m(x,y)} + e^{m(x,n+1)}}{\sum_{y'\in\bar{\mathcal{Y}}}e^{m(x,y')}}\biggr],
\end{equation}
where $\alpha$ is the cost of non-deferral, $m(x,y)$ corresponds to logits for the target class, and $m(x, n+1)$ to logits for the deferral class. Different deferral rates can be realised by retraining the method with varying cost terms $\alpha$.

\textbf{Learned deferral (two-stage)} -- Secondly, we examine the two-stage deferral approach proposed by Liu et al.~\cite{liu2022incorporating}. The first stage of this approach corresponds to an ensemble to estimate uncertainty. 
The second stage is a multilayer perceptron (MLP), which takes as input the softmax outputs of the ensemble along with diagnostic entropy and model entropy to predict a class label $\hat{y}$ or defer (see \cite{liu2022incorporating} for details). The MLP is trained with the surrogate loss proposed in~\cite{mozannar2020consistent}, which is given by
\begin{equation}
\label{eq:two-stage}
    l(m,x,y) = -log\biggl[\dfrac{e^{m(x,y)}}{\sum_{y'\in\bar{\mathcal{Y}}}e^{m(x,y')}}\biggr] - \beta~ log\biggl[\dfrac{e^{m(x,n+1)}}{\sum_{y'\in\bar{\mathcal{Y}}}e^{m(x,y')}}\biggr],
\end{equation}
where $\beta$ is the deferral cost and is, again, adjustable to control the deferral rate.

\subsubsection{Deferral through Uncertainty Quantification}
\label{sec:def_UQ}
Uncertainty estimates obtained using various uncertainty quantification methods can guide whether to make an autonomous decision or defer to a human expert by applying a threshold. We study five ways to estimate uncertainty.

\textbf{Softmax-based} -- A naive approach is to directly use the softmax as a proxy for the model's confidence. For binary classification, the model is most uncertain at softmax values close to 0.5 and confident at values close to zero and one. 
To obtain an uncertainty measure from the softmax outputs, we apply the projection function $u(x) = 1 - 2 \cdot |s_1 - 0.5|$ to the softmax outputs of the positive class $s_1$. This ensures that the uncertainty ranges from zero (certain) to one (uncertain). This approach requires no sampling.

\textbf{Ensembles}~\cite{lakshminarayanan2017simple} -- The ensemble method is implemented by training $N$ ResNet50s with different random seeds. Sampling is done by passing the input through each model in the ensemble. 

\textbf{SWAG}~\cite{maddox2019simple} -- This method is an extension of stochastic weight averaging (SWA)~\cite{izmailov2018averaging}, which computes the mean over stochastic gradient descent (SGD) iterates. SWAG then fits a Gaussian by using the SWA solution as the first momentum and a low rank plus diagonal covariance also derived from the SGD iterates, which captures the uncertainty of the model weights. Samples are obtained by sampling from the Gaussian distribution $N$ times.

\textbf{MC Dropout}~\cite{gal2016dropout} --  In this widely used method, dropout is not only applied during training but also during inference. By feeding the input through the model $N$ times with dropout enabled, the samples can be treated as Monte Carlo samples from the posterior over neural network weights. We applied dropout with a rate of $p=0.2$ after the final average pooling layer of a ResNet50. 

\textbf{Bayesian Neural Networks} -- We implement a Bayesian ResNet50 using the Flipout~\cite{wen2018flipout} technique, which is a refinement of Bayes by Backprop~\cite{blundell2015weight}.
This approach learns a Gaussian distribution over the model’s weights, resulting in a posterior distribution of the network’s parameters. Sampling is performed by passing the input through the model $N$ times.

We set $N = 10$ for all methods relying on sampling. The uncertainty is then obtained by computing the variance over the $N$ softmax outputs. During training, the minority class, Glaucoma, was oversampled with sampling weights determined by one divided by the number of images per class label. Model selection was done using pAUC (AUC at 90-100\% specificity) for Ensemble, Softmax, SWAG, MC Dropout, and BNN and minimum loss for the learned deferral models on the validation set.

\subsection{Data}
\label{sec:data}
We tested the above methods on the fundus image dataset from the Artificial Intelligence for Robust Glaucoma Screening (AIROGS) challenge~\cite{de2023airogs}. This dataset comprises 101'442 images that are highly diverse in appearance.
Moreover, with only about $3\%$ of images displaying Glaucoma, the dataset is highly imbalanced, thereby representing real-world clinical settings. We divided the data into train, validation, test splits at a ratio of 70/20/10. All images were preprocessed by cropping around the fundus to result in a square image and resizing to 320x320 using the methodology proposed in~\cite{gervelmeyer2025fundus}.

\subsection{Evaluation Metric}

Good deferral performance means that the model is able to identify those images that it will likely misclassify autonomously and instead defer them. We therefore measure deferral performance at a specific deferral rate using the classification performance on the remaining, non-deferred, images. If the correct images are deferred, classification performance should increase on the remaining images. 
We measure the classification performance using balanced accuracy (bAcc) with a decision threshold of 0.5, which we found to be a more reliable measure than AUC due to the severe class imbalance of the AIROGS dataset. 
We additionally report the fraction of positives cases that are deferred for each deferral rate.

\section{Experiments and Results}
\label{sec:experiments}

We evaluated the different deferral approaches in terms of classification performance, deferral of in-distribution (ID) data, and deferral of OOD data. 

\subsection{Classification Performance}

We first compared all methods in their ability to accurately predict glaucoma from fundus images. To evaluate the learned deferral methods, we used one model per method that did not defer any images under the chosen deferral cost. 
As can be seen in Tab.~\ref{tab:classification}, all models achieved good classification performance with the ensemble reaching the highest AUC of 0.852. The difference in performance on the positive Glaucoma class is noteworthy. For that class, most methods only reached an accuracy of around 0.5 with SWAG reaching best performance on the positive class with an accuracy of 0.705, followed by MC Dropout (0.646) and the one-stage learned deferral model (0.643).

\begin{table}[]
\caption{\textbf{Classification performance.} AUC, balanced accuracy (bAcc) and per-class accuracy (Acc$_0$, Acc$_1$) for each method.}
\label{tab:classification}
\begin{tabularx}{\textwidth}{lXXXX}
Method                       & AUC                       & bAcc                      & Acc$_0$                    & Acc$_1$                             \\
\toprule
Ensemble                     & \textbf{0.852}            & 0.735                     & 0.976                     & 0.494                              \\
Softmax                      & 0.810                     & 0.735                     & 0.938                     & 0.532                              \\
SWAG                         & 0.838                     & \textbf{0.766} & 0.848 & \textbf{0.705} \\
BNN                     & 0.799 & 0.674 & 0.958 & 0.390      \\
MC Dropout                   & 0.827                     & \textbf{0.766}            & 0.885                     & 0.646                    \\
Learned Deferral (one-stage) & 0.836                     & 0.746                     & 0.850                     & 0.643                              \\
Learned Deferral (two-stage) & 0.818                     & 0.711                     & \textbf{0.981}            & 0.442                             
\end{tabularx}

\end{table}

\subsection{In-distribution Deferral Performance}
\label{sec:id-deferral}

Next, we investigated deferral performance on the test set, which was obtained using a random split and can be considered ID. Specifically, we evaluated each of the models at different deferral rates. For the learned deferral models, we trained ten models each with different deferral costs $\alpha, \beta$ to obtain models with deferral rates spanning the full range from zero to one (refer to the code for details). For the uncertainty-based methods we computed data points at different deferral rates by progressively lowering the uncertainty threshold used for deferral in 200 steps. 

\begin{figure}[]
    \centering
    \includegraphics[width=\textwidth]{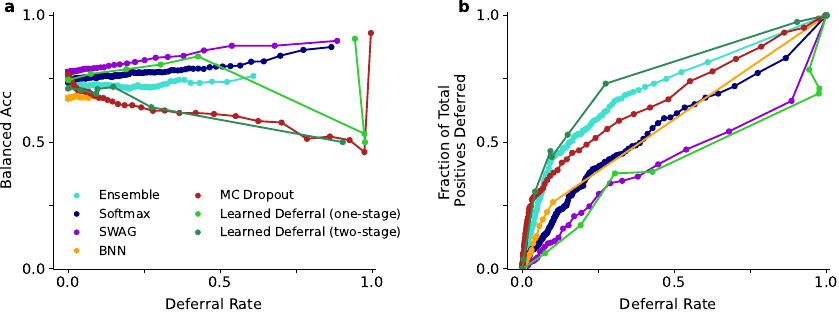}
    \caption{\textbf{In-distribution deferral performance.} a) bAcc of non-deferred images vs. deferral rate. Good deferral should lead to increasing classification performance on remainder of the dataset. b) Fraction of total positives for each deferral rate.}
    \label{fig:in-domain}
\end{figure}

As can be seen in Fig.~\ref{fig:in-domain}a, the learned deferral models both initially correctly deferred images with wrong predictions but quickly degraded in deferral performance at around $15\%$ or $40\%$ deferral, respectively. Another observation was that the models were very sensitive to the deferral cost parameters $\alpha, \beta$ with slight changes in those parameters leading to major differences in deferral rate. A consequence of this is that there are few data points for higher deferral rates as finding $\alpha, \beta$ parameters that would fall in this range proved very difficult. These limitations would make it difficult to apply the models in real clinical settings.
An interesting observation was that the two-stage deferral approach appeared to have learned an unwanted deferral short-cut. As can be seen in Fig.~\ref{fig:in-domain}b, it learned to predominately defer positive images regardless of its performance on those images, as is evidenced by the decreasing performance curve in Fig.~\ref{fig:in-domain}a. 

In contrast, the uncertainty-based Softmax, BNN and SWAG methods performed as expected with increasing bAcc with growing deferral rate, meaning they correctly deferred  images that they would otherwise misclassify. MC Dropout formed the exception with consistently decreasing accuracy with higher deferral rates. 
However, we also observed that the deferral rates for the uncertainty-based methods were more heavily concentrated near zero, indicating that all methods were overly confident. Moreover, small changes in the deferral threshold sometimes caused sudden jumps in the deferral rate leading to total deferral, for which no accuracy can be calculated. This explains why some curves in Fig.~\ref{fig:in-domain}b stop early. BNNs tended to produce particularly small uncertainty estimates, which made the method particularly sensitive to the threshold and caused the deferral rates to closely bunch together. This could be alleviated with even finer threshold spacing, however, we opted against this, as the threshold steps would fall into the 4th or 5th decimal.

\subsection{Out-of-distribution Deferral Performance}

An OOD setting was simulated by corrupting the test images in two different ways: corruption with five increasing levels of Gaussian noise and with five increasing levels of Gaussian blur (see Appendix for details). We then repeated the experiments from Section~\ref{sec:id-deferral} on the corrupted datasets. We also considered introducing realistic artifacts to create an OOD dataset. However, the AIROGS dataset is so comprehensive that most naturally occurring artifacts are in-fact in the training data. We therefore did not pursue this further.

As deferral behaviour was similar for all OOD levels we only show the results for the level 3 corruptions in Fig.~\ref{fig:ood}. The remaining levels can be found in the Appendix. We observed that the sensitivity of the learned deferral methods to the deferral cost parameters became more pronounced in the OOD setting. For the one-stage learned deferral approach decreasing the deferral cost sometimes even lead to an unexpected decrease in the referral rate as can be seen in Fig.~\ref{fig:ood} where the deferral rate reverses direction for high deferral rates (light green curve). The two-stage learned deferral approach did not show this behaviour, however, it did not perform well in the OOD setting in particular for the noise corruption where the approach exhibited decreasing performance for each deferral rate, i.e. it is consistently deferring correctly classified images rather than wrongly classified ones. 
Moreover, the two-stage method showed the same short-cut learning behaviour observed in the ID experiments (see deferral fraction plots in the Appendix). We conclude that, as hypothesised, learned deferral methods struggle in the OOD setting limiting their practical applicability.

Again, the uncertainty-based methods showed a better deferral behaviour with SWAG leading to the best results and MC Dropout performing the worst.   
Overall, we observed that the blurring is less impactful on classification and deferral performance with most models behaving similarly to their performance on ID data. In contrast, performance on noisy data degrades more noticeably. Only the SWAG and BNN models continue to perform as expected, although it's important to note that BNNs again estimate very small uncertainty values, which would make threshold tuning impractical in practice.

\label{sec:ood}
\begin{figure}[]
    \centering
    \includegraphics[width=\textwidth]{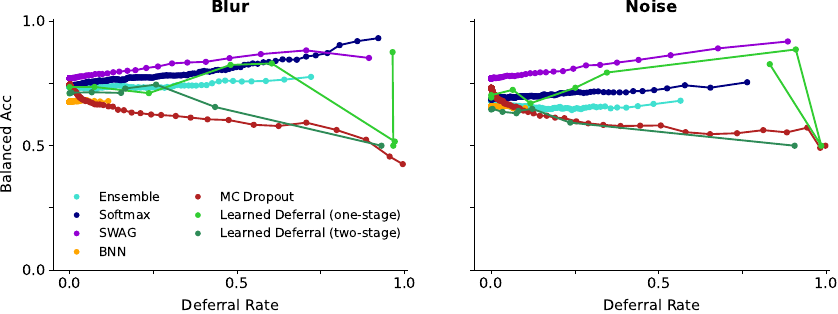}
    \caption{\textbf{Out-of-distribution (OOD) deferral.} Deferral performance on OOD images corrupted by a moderate level of  Gaussian blurring and Gaussian noise.}
    \label{fig:ood}
\end{figure}
We also evaluated the OOD experiment on real OOD data, namely fundus images from the Chákṣu~\cite{kumar2023chakṣu} dataset captured with a device that was not used in AIROGS. However, we found that none of the models achieved acceptable classification performance, which would make further deferral performance investigations meaningless (see Appendix). This result further highlights the extreme sensitivity of current AI models to distributional shifts.

\section{Discussion and Conclusion}
In this work, we set out to investigate whether uncertainty quantification methods could provide a viable alternative to learned deferral approaches on clinically realistic ophthalmology data. Our experiments revealed a number of shortcomings of learned deferral methods that could limit their usefulness in practice. Firstly, the two investigated learned deferral methods did not show good deferral behaviour especially when compared to some of the methods based on uncertainty quantification. Secondly, learned deferral methods, by design, rely on the training objective and are therefore sensitive to the choice of cost function. As we show, this can lead to shortcut learning based on unintended confounders in the data, such as developing a deferral bias toward a specific class label. Lastly, learned deferral approaches proved to be more sensitive to OOD data, which severely limits their practical applicability in clinical settings. 

In contrast we found that uncertainty quantification methods provide a promising alternative to the learning to defer paradigm. Their deferral performance was generally better, with SWAG being a particularly promising approach. Furthermore, methods based on uncertainty quantification are not sensitive to the deferral cost and are not prone to learning shortcuts because the model cannot foresee during training how its uncertainty estimate will ultimately be used. 

While our study produced promising evidence for using uncertainty quantification methods for deferral over learning to defer approaches, the study has several limitations. Firstly, we only evaluated two learning to defer methods. While we believe that they are representative of the general limitations of this family of approaches, in future work we plan to evaluate more methods from this category. Secondly, the evaluation was only performed on a single dataset. In the future, we aim to extend the study to other clinically relevant AI deferral scenarios such as emergency x-ray scans. 

\begin{credits}
\subsubsection{\ackname} Funded by the Deutsche Forschungsgemeinschaft (DFG) – EXC number 2064/1 – Project number 390727645.

\subsubsection{\discintname}
The authors have no competing interests to declare that are relevant to the content of this article. 
\end{credits}
 \bibliographystyle{splncs04}
 \bibliography{bibliography}

\begin{thebibliography}{10}
\providecommand{\url}[1]{\texttt{#1}}
\providecommand{\urlprefix}{URL }
\providecommand{\doi}[1]{https://doi.org/#1}

\bibitem{alves2024cost}
Alves, J.V., Leit{\~a}o, D., Jesus, S., Sampaio, M.O., Li{\'e}bana, J., Saleiro, P., Figueiredo, M.A., Bizarro, P.: Cost-sensitive learning to defer to multiple experts with workload constraints. arXiv preprint arXiv:2403.06906  (2024)

\bibitem{beede2020human}
Beede, E., Baylor, E., Hersch, F., Iurchenko, A., Wilcox, L., Ruamviboonsuk, P., Vardoulakis, L.M.: A human-centered evaluation of a deep learning system deployed in clinics for the detection of diabetic retinopathy. In: Proceedings of the 2020 CHI conference on human factors in computing systems. pp. 1--12 (2020)

\bibitem{blundell2015weight}
Blundell, C., Cornebise, J., Kavukcuoglu, K., Wierstra, D.: Weight uncertainty in neural network. In: International conference on machine learning. pp. 1613--1622. PMLR (2015)

\bibitem{de2024towards}
De~Toni, G., Okati, N., Thejaswi, S., Straitouri, E., Gomez-Rodriguez, M.: Towards human-ai complementarity with predictions sets. arXiv preprint arXiv:2405.17544  (2024)

\bibitem{de2023airogs}
De~Vente, C., Vermeer, K.A., Jaccard, N., Wang, H., Sun, H., Khader, F., Truhn, D., Aimyshev, T., Zhanibekuly, Y., Le, T.D., et~al.: Airogs: Artificial intelligence for robust glaucoma screening challenge. IEEE transactions on medical imaging  \textbf{43}(1),  542--557 (2023)

\bibitem{dvijotham2023enhancing}
Dvijotham, K., Winkens, J., Barsbey, M., Ghaisas, S., Stanforth, R., Pawlowski, N., Strachan, P., Ahmed, Z., Azizi, S., Bachrach, Y., et~al.: Enhancing the reliability and accuracy of ai-enabled diagnosis via complementarity-driven deferral to clinicians. Nature Medicine  \textbf{29}(7),  1814--1820 (2023)

\bibitem{fang2024learning}
Fang, Y., Nalisnick, E.: Learning to defer with an uncertain rejector via conformal prediction. In: NeurIPS 2024 Workshop on Bayesian Decision-making and Uncertainty

\bibitem{gal2016dropout}
Gal, Y., Ghahramani, Z.: Dropout as a bayesian approximation: Representing model uncertainty in deep learning. In: international conference on machine learning. pp. 1050--1059. PMLR (2016)

\bibitem{gervelmeyer2025fundus}
Gervelmeyer, J., M{\"u}ller, S., Huang, Z., Berens, P.: Fundus image toolbox: A python package for fundus image processing. Journal of Open Source Software  \textbf{10}(108), ~7101 (2025)

\bibitem{he2016deep}
He, K., Zhang, X., Ren, S., Sun, J.: Deep residual learning for image recognition. In: Proceedings of the IEEE conference on computer vision and pattern recognition. pp. 770--778 (2016)

\bibitem{hemmer2022forming}
Hemmer, P., Schellhammer, S., Vössing, M., Jakubik, J., Satzger, G.: Forming effective human-ai teams: building machine learning models that complement the capabilities of multiple experts. arXiv preprint arXiv:2206.07948  (2022)

\bibitem{heydon2021prospective}
Heydon, P., Egan, C., Bolter, L., Chambers, R., Anderson, J., Aldington, S., Stratton, I.M., Scanlon, P.H., Webster, L., Mann, S., et~al.: Prospective evaluation of an artificial intelligence-enabled algorithm for automated diabetic retinopathy screening of 30 000 patients. British Journal of Ophthalmology  \textbf{105}(5),  723--728 (2021)

\bibitem{izmailov2018averaging}
Izmailov, P., Podoprikhin, D., Garipov, T., Vetrov, D., Wilson, A.G.: Averaging weights leads to wider optima and better generalization. arXiv preprint arXiv:1803.05407  (2018)

\bibitem{keswani2021towards}
Keswani, V., Lease, M., Kenthapadi, K.: Towards unbiased and accurate deferral to multiple experts. In: Proceedings of the 2021 AAAI/ACM Conference on AI, Ethics, and Society. pp. 154--165 (2021)

\bibitem{kumar2023chakṣu}
Kumar, J.H., Seelamantula, C.S., Gagan, J., Kamath, Y.S., Kuzhuppilly, N.I., Vivekanand, U., Gupta, P., Patil, S.: Ch{\'a}kṣu: A glaucoma specific fundus image database. Scientific data  \textbf{10}(1), ~70 (2023)

\bibitem{lakshminarayanan2017simple}
Lakshminarayanan, B., Pritzel, A., Blundell, C.: Simple and scalable predictive uncertainty estimation using deep ensembles. Advances in neural information processing systems  \textbf{30} (2017)

\bibitem{lambert2022trustworthy}
Lambert, B., Forbes, F., Tucholka, A., Doyle, S., Dehaene, H., Dojat, M.: Trustworthy clinical ai solutions: a unified review of uncertainty quantification in deep learning models for medical image analysis. arXiv preprint arXiv:2210.03736  (2022)

\bibitem{liu2022incorporating}
Liu, J., Gallego, B., Barbieri, S.: Incorporating uncertainty in learning to defer algorithms for safe computer-aided diagnosis. Scientific reports  \textbf{12}(1), ~1762 (2022)

\bibitem{macdonald2025generating}
Macdonald, T., Zhelev, Z., Liu, X., Hyde, C., Fajtl, J., Egan, C., Tufail, A., Rudnicka, A.R., Shinkins, B., Given-Wilson, R., et~al.: Generating evidence to support the role of ai in diabetic eye screening: considerations from the uk national screening committee. The Lancet Digital Health  \textbf{7}(5) (2025)

\bibitem{maddox2019simple}
Maddox, W.J., Izmailov, P., Garipov, T., Vetrov, D.P., Wilson, A.G.: A simple baseline for bayesian uncertainty in deep learning. Advances in neural information processing systems  \textbf{32} (2019)

\bibitem{mao2024two}
Mao, A., Mohri, C., Mohri, M., Zhong, Y.: Two-stage learning to defer with multiple experts. Advances in neural information processing systems  \textbf{36} (2024)

\bibitem{mao2024predictor}
Mao, A., Mohri, M., Zhong, Y.: Predictor-rejector multi-class abstention: Theoretical analysis and algorithms. In: International Conference on Algorithmic Learning Theory. pp. 822--867. PMLR (2024)

\bibitem{mao2024principled}
Mao, A., Mohri, M., Zhong, Y.: Principled approaches for learning to defer with multiple experts. In: International Workshop on Combinatorial Image Analysis. pp. 107--135. Springer (2024)

\bibitem{mao2024realizable}
Mao, A., Mohri, M., Zhong, Y.: Realizable $ h $-consistent and bayes-consistent loss functions for learning to defer. arXiv preprint arXiv:2407.13732  (2024)

\bibitem{mozannar2023should}
Mozannar, H., Lang, H., Wei, D., Sattigeri, P., Das, S., Sontag, D.: Who should predict? exact algorithms for learning to defer to humans. In: International conference on artificial intelligence and statistics. pp. 10520--10545. PMLR (2023)

\bibitem{mozannar2020consistent}
Mozannar, H., Sontag, D.: Consistent estimators for learning to defer to an expert. In: International conference on machine learning. pp. 7076--7087. PMLR (2020)

\bibitem{ruamviboonsuk2022real}
Ruamviboonsuk, P., Tiwari, R., Sayres, R., Nganthavee, V., Hemarat, K., Kongprayoon, A., Raman, R., Levinstein, B., Liu, Y., Schaekermann, M., et~al.: Real-time diabetic retinopathy screening by deep learning in a multisite national screening programme: a prospective interventional cohort study. The Lancet Digital Health  \textbf{4}(4),  e235--e244 (2022)

\bibitem{verma2022calibrated}
Verma, R., Nalisnick, E.: Calibrated learning to defer with one-vs-all classifiers. In: International Conference on Machine Learning. pp. 22184--22202. PMLR (2022)

\bibitem{wen2018flipout}
Wen, Y., Vicol, P., Ba, J., Tran, D., Grosse, R.: Flipout: Efficient pseudo-independent weight perturbations on mini-batches. arXiv preprint arXiv:1803.04386  (2018)

\bibitem{zhang2025learning}
Zhang, Z., Ai, W., Wells, K., Rosewarne, D., Do, T.T., Carneiro, G.: Learning to complement and to defer to multiple users. In: European Conference on Computer Vision. pp. 144--162. Springer (2025)

\end{thebibliography}
 
 \clearpage
\appendix
\section*{Supplementary Materials}

\renewcommand{\thesubsection}{\Alph{subsection}}

\section{Out-of-distribution Simulation through Corruption}
\begin{figure}[]
    \centering
    \includegraphics[width=\textwidth]{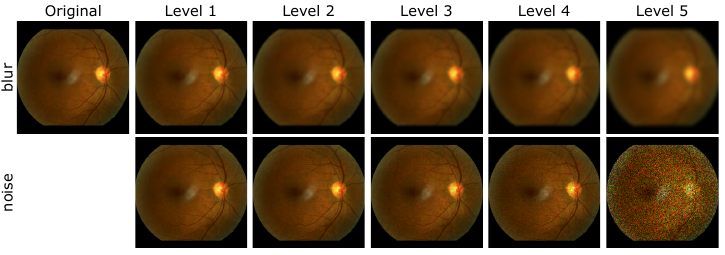}
    \caption{\textbf{Corruption of images.} We investigated five different levels of corruption (Gaussian blurring and noise) and repeated the corruption process on the whole test dataset to ensure that each level includes the same images.}
    \label{apx:fig:oods}
\end{figure}
\section{Deferral Performance in OOD}
\begin{figure}[]
    \centering
    \includegraphics[width=\textwidth]{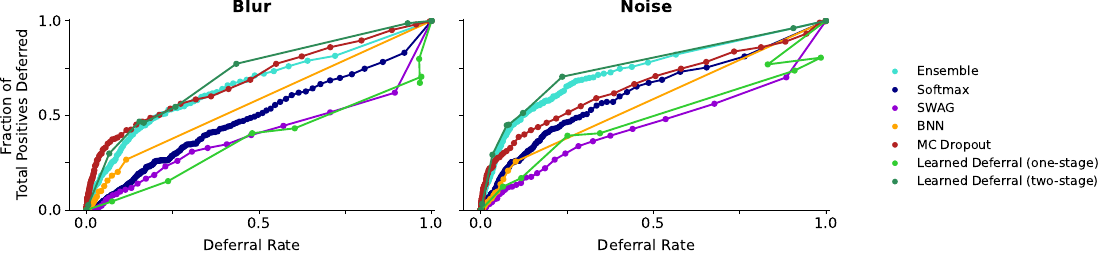}
    \caption{\textbf{Deferral of positives on moderately corrupted images (level 3).} }  
    \label{apx:fig:ood-positives}
\end{figure}
\begin{figure}[]
    \centering
    \includegraphics[width=0.7\textwidth]{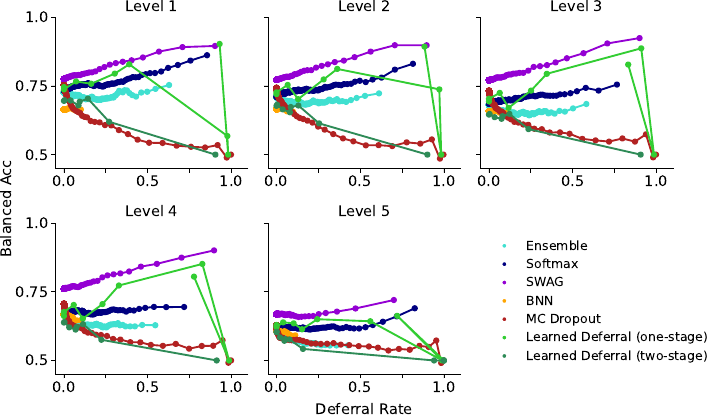}
    \caption{\textbf{OOD deferral per level of Gaussian noise corruption.} }  
    \label{apx:fig:ood-noise}
\end{figure}

\begin{figure}[]
    \centering
    \includegraphics[width=0.7\textwidth]{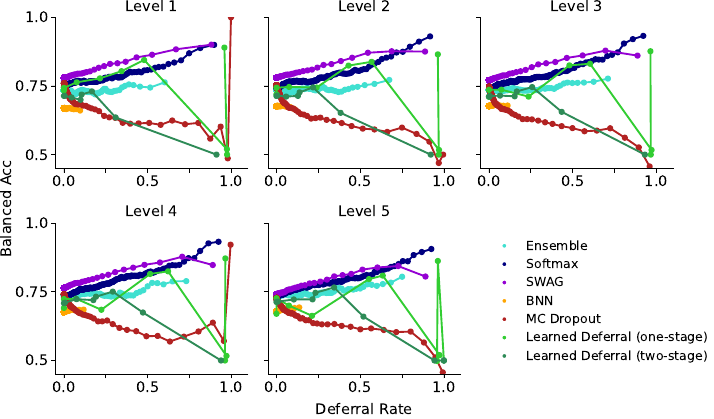}
    \caption{\textbf{OOD deferral per level of Gaussian blur corruption.} }  
    \label{apx:fig:ood-blur}
\end{figure}

\section{Classification Performance on Chákṣu}
\begin{table}[]
\label{apx:chaksu_classification}
\caption{\textbf{Classification performance on Chákṣu.} AUC, balanced accuracy (bAcc) and per-class accuracy (Acc$_0$, Acc$_1$) for each method.}
\label{tab:classification}
\begin{tabularx}{\textwidth}{lXXXX}
Method                       & AUC                       & bAcc                      & Acc$_0$                    & Acc$_1$                             \\
\toprule
Ensemble                     & 0.532            & 0.508                     & 0.974                     & 0.042                              \\
Softmax                      & 0.411                     & 0.444                     & 0.846                     & 0.042                              \\
SWAG                         & 0.583                     & 0.461 & 0.214 & \textbf{0.708} \\
BNN                     & \textbf{0.606} & 0.541 & 0.915 & 0.167      \\
MC Dropout                   & 0.575                     & \textbf{0.552}            & 0.812                     & 0.292                    \\
Learned Deferral (one-stage) & 0.567                     & 0.548                     & 0.513                     & 0.583                              \\
Learned Deferral (two-stage) & 0.569                     & 0.496                     & \textbf{0.991}            & 0.000                             
\end{tabularx}
\end{table}

\end{document}